\documentclass[sigconf]{acmart}
\usepackage{algorithm, algorithmic}
\usepackage{multirow}
\usepackage{enumitem} 
\AtBeginDocument{%
  \providecommand\BibTeX{{%
    \normalfont B\kern-0.5em{\scshape i\kern-0.25em b}\kern-0.8em\TeX}}}

\setcopyright{acmcopyright}
\copyrightyear{2022}
\acmYear{2022}
\acmDOI{10.1145/3534678.3539159}

\acmConference[KDD '22]{KDD '22: ACM SIGKDD Conference on Knowledge Discovery and Data Mining}{August 14--18, 2022}{Washington DC}
\acmBooktitle{KDD '22: ACM SIGKDD Conference on Knowledge Discovery and Data Mining,
  August 14--18, 2022, Washington DC}
\acmPrice{15.00}
\acmISBN{978-1-4503-9385-0/22/08}

\begin{document}
\title{Temporal Multimodal Multivariate Learning}

\author{Hyoshin Park}
\email{hpark1@ncat.edu}
\orcid{0002-1490-5404}
\affiliation{%
  \institution{North Carolina A\&T State University\\Dept. of Comput. Data Sci. \& Eng.}
  \streetaddress{}
  \city{}
  \state{}
  \country{}
  \postcode{}
}

\author{Justice Darko}
\authornote{Both authors contributed equally to this research.}
\author{Niharika Deshpande}
\authornotemark[1]
\affiliation{%
  \institution{North Carolina A\&T State University\\Dept. of Comput. Data Sci. \& Eng.}
  \streetaddress{}
  \city{}
  \state{}
  \country{}
  \postcode{}
\email{jdarko@aggies.ncat.edu}}

\author{Venktesh Pandey}
\affiliation{%
  \institution{North Carolina A\&T State University\\Dept. of Civil, Archi., \& Envir.  Eng.}
  \streetaddress{}
  \city{Greensboro, NC}
  \country{USA}
\email{vpandey@ncat.edu}}

\author{Hui Su}
\affiliation{%
  \institution{Stratosphere \& Upper Troposphere \\Jet Propulsion Laboratory\\ CalTech, Pasadena, CA, USA}
  \streetaddress{}
  \country{}}
\email{hui.su@jpl.nasa.gov}

\author{Masahiro Ono}
\affiliation{%
  \institution{Robotics \\Jet Propulsion Laboratory\\ CalTech, Pasadena, CA, USA}
  \streetaddress{}
  \country{}}
\email{masahiro.ono@jpl.nasa.gov}

\author{Dedrick Barkely}
\authornote{Both authors contributed equally to this research.}
\author{Larkin Folsom}
\authornotemark[2]
\affiliation{%
  \institution{North Carolina A\&T State University\\Dept. of Comput. Data Sci. \& Eng.}
  \streetaddress{}
  \city{Greensboro, NC}
  \state{}
  \country{USA}
  \postcode{}}

\author{Derek Posselt}
\affiliation{%
 \institution{Atmospheric Physics \& Weather \\Jet Propulsion Laboratory, California Institute of Technology, Pasadena}
  \streetaddress{}
  \country{}}
\email{derek.posselt@jpl.nasa.gov}

\author{Steve Chien}
\affiliation{%
  \institution{Artificial Intelligence \\Jet Propulsion Laboratory, California Institute of Technology, Pasadena}
  \streetaddress{}
  \country{}}
\email{steve.a.chien@jpl.nasa.gov}

\renewcommand{\shortauthors}{Hyoshin Park, et al.}
\begin{abstract}
  We introduce temporal multimodal multivariate learning, a new family of decision making models that can indirectly learn and transfer online information from simultaneous observations of a probability distribution with more than one peak or more than one outcome variable from one time stage to another. We approximate the posterior by sequentially removing additional uncertainties across different variables and time, based on data-physics driven correlation, to address a broader class of challenging time-dependent decision-making problems under uncertainty. Extensive experiments on real-world datasets ( i.e., urban traffic data and hurricane ensemble forecasting data) demonstrate the superior performance of the proposed targeted decision-making over the state-of-the-art baseline prediction methods across various settings.
\end{abstract} 

\begin{teaserfigure}
  \includegraphics[width=\textwidth]{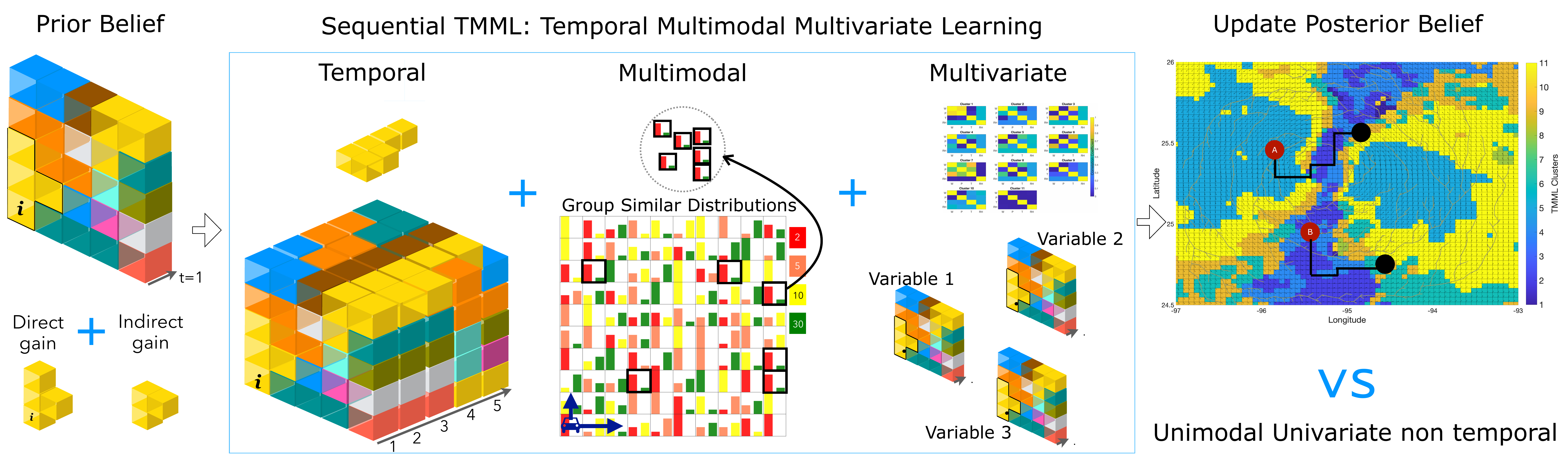}
  \caption{Online gain in temporal, multimodal, and multivariate prediction uncertainty between prior and posterior. Each cell can be assumed to have a combination of discrete travel time distributions (i.e., 2, 5, 10, 30min) with different weights.}
  \Description{}
  \label{fig:teaser}
\end{teaserfigure}

\maketitle
\section{Introduction}
In recent years, deep reinforcement learning (RL) models have improved the solution quality of online combinatorial optimization problems \cite{Bello17, Hanjun17}, yet cannot match the real-world online systems \cite{Gauci19}. Consider a Mars autonomous navigation problem under uncertainty where information sampled from aerial agents mapping large areas and ground agents observing traversability can be transferred to other agents for safe and efficient navigation \cite{ono2020maars}. For the posterior approximation \cite{BENGIO2021405} (e.g., Markov decision process) with online data assimilation \cite{folsom2020}, we need a new framework to actively sample useful information. Traditional information metrics like Shannon Entropy or Kullback-Leibler (KL) Divergence fail to incorporate future uncertainty with more than one peak probability distributions \cite{folsom2020}. Shannon Entropy \cite{shannon1949} cannot distinguish distributions with multiple weights (e.g., bimodal distributions) because it only considers raw information gain, treating all information as equally valuable. KL Divergence \cite{kullback1951} introduces a bias toward only one mode (e.g., Exclusive, Reverse) or toward the mean of the modes (e.g. Inclusive, Forward) with non-symmetrical measures of information gain. Recent computer vision models \cite{gupta2015kldiv} cannot address unobserved heterogeneity causing multimodal distributions since representing the information gain using KL Divergence requires comparison to an ``ideal'' distribution. This biases the model toward searching only for some types of uncertain solutions while ignoring other potentially more valuable solutions. When the probability distribution is heavily weighted at either extreme, the system cost either experience very high true savings or negative true savings. Those combinations of probability distributions vary across time and location and evolve as new observations become available.

Our main contribution is the development of a new family of online predictive decision making models, Temporal Multimodal Multivariate Learning (TMML), that can indirectly learn and transfer online information from multiple modes of probability distributions and multiple variables between different time stages, which can be applied to many routing problems under uncertainty such as Mars exploration \cite{folsom2020}, Hurricane sensing \cite{parkESS2020}, and urban routing \cite{folsomdynamic}. Preliminary remedy \cite{folsom2020} partially filled this gap by grouping similar types of locations based on their classified output (e.g., sandy or rocky), used in optimizing vehicle routing to improve the prediction uncertainty proven to be superior to partially observable Markov decision processes. Locations with broad bimodal distributions offered the greatest potential delta between the expected and true savings. We expand this bimodal learning to multimodal learning and the maximum information gain is accomplished by identifying the time-dependent similarity between the probability distribution of variables. With existing routing algorithms, opportunities for data collection are commonly skipped or missed entirely. A technology to collect more valuable observations while carefully spending system resources will add significant value to the autonomous decisions. The result will be an increased likelihood of encountering unexpected scientific discoveries, creating new opportunities to characterize uncertainties, reconciling the desire to explore further with the desire to explore in-depth, and eliminating the dichotomy between engineering limitations and scientific discovery.

Cells in the grid of Figure \ref{fig:teaser} with a similar combination of distributions are clustered together based on the similarity between the combinations (e.g., 6 cells outlined in black). As users traverse the map, exploration of a cell in a cluster will remove the travel time uncertainty of other cells in the same cluster. In other words, exploring one cell of the cluster will identify which of the two travel time distributions applies to that explored cell, and to all other unexplored cells in the cluster. However, each cell has two travel time distributions with peaks of different heights. Therefore, we do not update all cells that share a single travel time distribution; we update cells that have similar combinations of distributions. This technique can be applied to several real-life applications. For example, assume that each cell with heterogeneous users presents a mixture of traffic conditions \cite{PARK2010662}. An online RL simplifies multi-modality to a unimodal distribution $ X \sim \mathcal{N}(35,\,10^{2})$ resulting in a lost opportunity to remove uncertainties in other locations. 

Several techniques learn and transfer information gained from multimodal distribution data in information theory \cite{shannon1949, kullback1951} for global uncertainty removal: grouping similar combinations of distributions, sampling from similar groups and updating posteriors, and solving probabilistic optimization \cite{Bello17, BENGIO2021405} for online routing. Those are necessary to optimize the probabilistic global routing problems based on knowledge learned and transferred in a sequence, and data is typically obtained from parts and not analyzed as a whole. While previous research \cite{folsom2020, folsom2021, parkESS2020} addressed bimodal learning and full uncertainty removal, we address multimodal learning integrated with partial information gains from temporal and multivariate learning applied to urban traffic and hurricane data. 
\vspace{-0.3em}
\section{Multimodal Learning}
\begin{figure}[!htb]
	\centering
	\includegraphics[width=\linewidth]{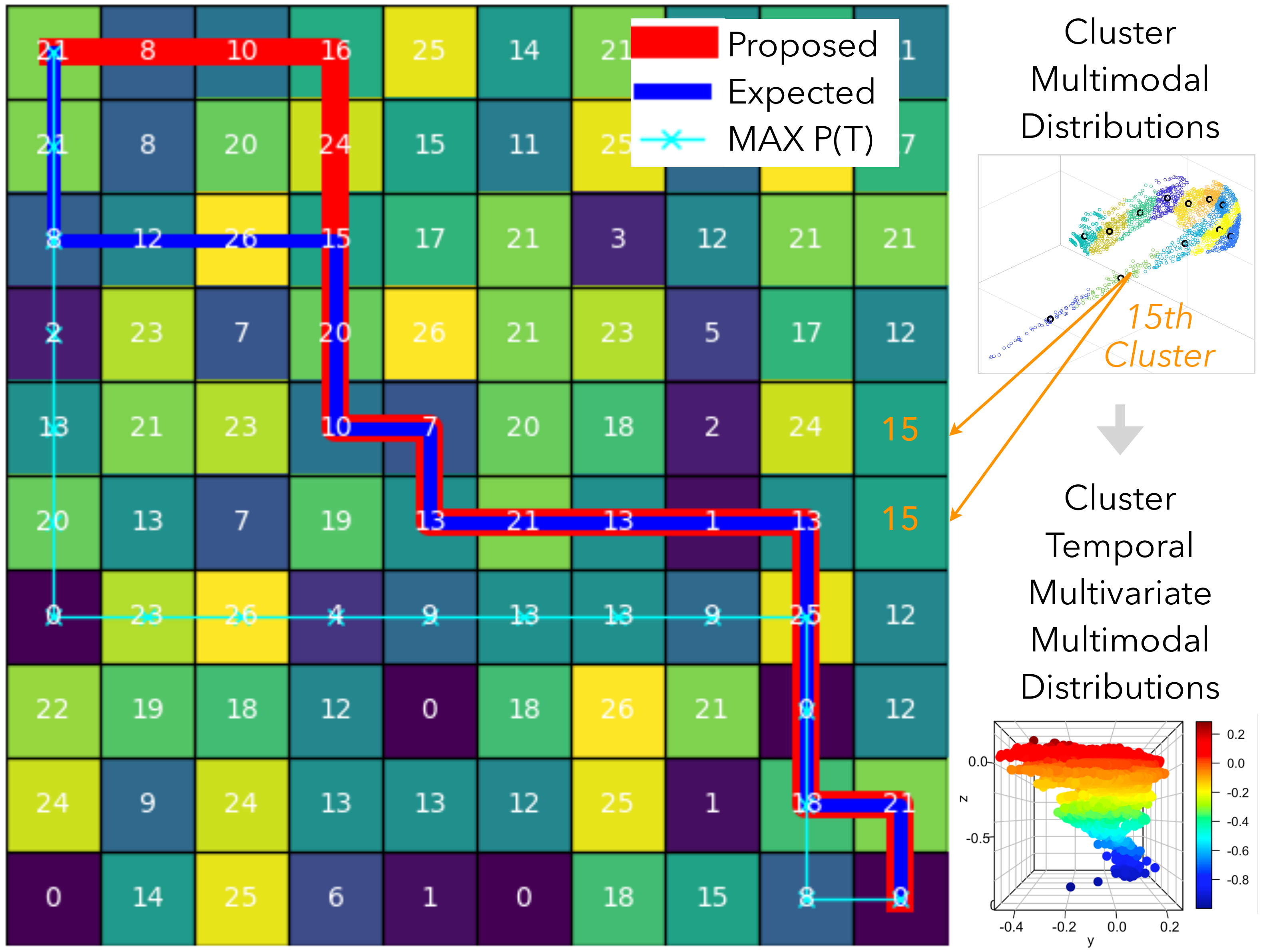}
	\caption{Multimodal extensions to the \textcolor{red}{proposed path}  \cite{folsom2020}, compared against theoretically  minimizing the initial expected travel time (\textbf{\textcolor{blue}{ETT}}) or the highest probability classification-based travel time the \textcolor{cyan}{Max$[P(T)]$}. The grid shows each cell type by number (white) and filled color.}
	\label{ttpgrida}\vspace{-1.5em}
\end{figure}

Reduced uncertainty in bimodal travel time information can be processed and transferred from one agent to another agent \cite{folsom2020}. A prototype of bimodal uncertainty removal in an \texttt{10}$\times$\texttt{10} grid map is extended to multimodality of each cell through clustering similar probability distributions for multimodal learning (Figure \ref{ttpgrida}). The agent is allowed to move in four directions: up, down, left, and right. Diagonal moves are not allowed. Cells in the grid are numbered row-wise, starting with zero for the first cell. Grids are used because the prior state of the map will be defined using image analysis, which defines the state of a region using pixels of a fixed size. Those 2, 5, 10, and 30 minutes from Figure \ref{fig:teaser} bimodal is expressed as clusters. 
The key statistics for travel time distribution in each cell are based on lower and upper bounds, with probability $P(T)$ represented as real numbers between 0 and 1 in the model \cite{folsom2020} in Figure \ref{ttpgrida} further extended to incorporate multimodality. 

In this study, multimodal learning enhances the scientific and engineering value of autonomous vehicles by finding the best routes based on the desired level of exploration, risk, and constraints. In the proposed exploration framework, each grid cell (Figure \ref{ttpgrida}) contains a unique probabilistic distribution of travel time for formulating the best options to travel with partial, sequential, and mixture of information gain, with various probability distributions. An example application is the Machine learning-based Analytics for Rover Systems (MAARS) \cite{ono2020maars}, where agents analyze images for autonomous driving feature detection, and assist scientists by selectively collecting data without interrupting drives. When agents travel through a grid map, information can be gained by visiting cells classified with uncertainty, observing the conditions in those cells, and estimating the true state of other cells and observations from other agents. Existing work on energy-optimal path planning \cite{Tompkins2006MissionlevelPP,doi:10.1177/0278364904046632,7059362} assume that energy consumption and generation are given or immediately derivable from an existing height map. However, without prediction of the agent's energy consumption and generation, these methods are myopic, simplified, and not a realistic optimization approach. An agent’s energy consumption and generation depend on interrelated factors such as terramechanics, the agent’s dynamics and kinematics, and terrain topography.  

\section{Multivariate multimodal learning}
Traditional machine learning frameworks overlook simultaneous observations of more than one outcome variable in different locations and times without lowering the prediction errors. Real-life data, behaviors, and problems (referring to objects, values, and attributes) are non-independent and non-identically distributed, whereas most analytical methods assume independent and identically distributed (IID) random variables. Unfortunately, the interdependent event relationship has been overlooked and future posterior events have been assumed independent from other events and systems. The dynamic impact area of a prior event could predict the probability of posterior events \cite{park16b, Park2018}. However, when frequent minor events are occurring in a sequence, due to high uncertainty, the literature could not reliably predict the dynamic spatiotemporal evolution of a mutual relationship between events \cite{park16c}. Machine learning with rule extraction \citep{park15a} partially alleviates Black box issues, but without an effort to reduce uncertainty by observing a ground truth, the routing solutions are still unreliable and intractable. In this paper, those dependencies are partially addressed by clustering multidimensional correlation data from multiple variables through deep clustering and when one cluster is updated, other variable data from the same cluster are also updated. 

In our case study, the multivariate clustering applied on hurricane Small Unmanned Aerial Systems (sUAS) observation includes four variables in each cell: wind speed, temperature, relative humidity, and pressure observed from the boundary layer. Each cell represents a specific location in the hurricane. Utilizing data manipulation techniques, we transform each variable to a single vector and combine each of the four vectors to create a multidimensional data matrix. By aggregating all cells in the map and grouping similar types of probability distributions of multiple variables, when we observe those variables at one location, uncertainties of other correlated variables at other locations are realized in this study. 


\section{Online Learning framework}
Temporal learning addresses the time-dependent realization of uncertainties of other correlated variables at other locations at other time stages, when we observe variables at one-time stage. The online learning framework updates sequential information based on the rapidly exploring random tree star (RRT*) algorithm. The RRT* algorithm finds an initial path solution based on an originally developed utility map of the environment conditioned on some constraint. As the agent follows this path of connected way-points (nodes) and makes new observations, the utility map is sequentially updated, generating recourse actions to accommodate the new information. Specifically, in each time stage, the utilities of the cells in the map are updated based on observations made at the agent's current location. At the first successor node from the agent's current location, the updated utility map is used to find a new node in a defined search region centered at the agent's current location. The search regions' radius is equal to the length of the edge connecting the current location and the first successor node. 

Considering the nodes in this region, we evaluate their utilities and select the node with the highest utility, replacing the initial first successor node from the current location. After pruning the previous edge connection, we then rewire the current location node to the new node.The new node is also rewired to the current location's second successor node. We repeat pruning and rewiring as the agent moves through the nodes and receives new information (updated utilities) until it reaches the target location. 

\vspace{-0.7em}
 \begin{algorithm}[!htpb]
	\caption{\textsf{:TMML-RRT* with Online Recourse}}\label{rrtstarett}
	\begin{algorithmic}
		\STATE T $\gets$ InitializeTree()\\
		\STATE T $\gets$ InsertNode($\emptyset$,$z_{init}$,T)\\
		\FOR{i=0 to i=N}
		\STATE $z_{rand} \gets$ Sample(i)\\
		\STATE $z_{nearest} \gets$ NearestTMML(T,$z_{rand}$)\\
		\STATE $(z_{new},U_{new}) \gets$ Steer($z_{nearest}, z_{rand}$)\\
		\IF{NoExceed($z_{new}$)}
		\STATE $z_{near} \gets$ Near($T,z_{new},|V|$) \\
		\STATE $z_{max} \gets$ ChooseParentTMML($z_{near},z_{nearest},z_{new}$)\\
		\STATE T $\gets$ InsertNode($z_{max},z_{new},T$)\\
		\STATE T $\gets$ Rewire($T,z_{near},z_{max},z_{new}$)\\
		\ENDIF
		\ENDFOR
		\STATE OptPath $\gets$ OnlineRecourse(T, $n_{s}, n_{g}$)\\
	\end{algorithmic}
\end{algorithm}
 The path search in Algorithm \ref{rrtstarett} shows the Nearest function in TMML-RRT* (NearestTMML) considering the utility at the nearest node. The NoExceed conditional statement implements constraints. The ChooseParent function considers the node with maximum utility within a defined region. The Rewire reevaluates previous connections from the agent’s start location and extends the new node to the node that can be accessed through the maximum utility. This process in Algorithm \ref{rrtstarett} is repeated until we find the target location. As shown in the pseudocode (Algorithm \ref{recourse1}) in Appendix \ref{appendixb}, after the initial path solution (based on the originally developed utility map) is found, the OnlineRecourse function is applied to find alternative waypoints (nodes) as the agent follows this path.

As the vehicle traverses its planned path, observations are made of the environment. Variational Bayesian inference generates a posterior belief given the prior belief of cell type distributions within each of the clusters. To measure how well variational multinomial posterior distribution $q_\lambda(z|x)$ approximates the true posterior $p(z|x)$, the $\mathbb{K}\mathbb{L}$ divergence $\mathbb{K}\mathbb{L}(q_\lambda(z|x)||p(z|x))$ estimates the information lost minimized with optimal variational parameters $\lambda$. The belief about the properties of different cell type clusters is updated en route to improve the travel. Clustering is performed using an expectation maximization algorithm on multinomial mixture models of the cells to identify cells with similar probability distributions. The likelihood of observing the dataset $P(\Upsilon|\mathbf{\alpha}, \mathbf{\beta})$ for data $\Upsilon$ and Dirichlet parameters $\mathbf{\alpha}$, $\mathbf{\beta}$ is the sum of $\alpha_k P(x_i | \beta_k)$ as observations $i$ goes from 1 to $N$ and clusters $k$ goes from 1 to $K$. Using Expectation Maximization, the optimal distribution of the data over $K$ clusters is determined by maximizing the lower bound of the log of the likelihood. The optimal cluster index minimizes the Bayesian Information Criterion as the difference between $D\ln{(N)}$ and $2 \ln{(\widehat{L})}$ where $D$ is the number of parameters, $N$ is the total number of observations, and $\widehat{L}$ is the likelihood of the model.

$D_{KL}$ quantifies the information gained by revising belief from the prior probability distribution $\mathcal{Q}$ to the posterior probability distribution $\mathcal{P}$. The mutual information $I(\mathcal{X};\mathcal{Y})$ between two discrete random variables $\mathcal{X}$ and $\mathcal{Y}$ is  $I(\mathcal{X};\mathcal{Y}) = D_{KL}(P_{(\mathcal{X},\mathcal{Y})}||P_\mathcal{X} \otimes P_{\mathcal{Y}})$
where $P_{(\mathcal{X},\mathcal{Y})}$ is the joint probability mass function and $P_{\mathcal{X}}$ and $P_{\mathcal{Y}}$ are the marginal probability mass functions of the random varaibles $\mathcal{X}$ and $\mathcal{Y}$. If $\mathcal{X}$ and $\mathcal{Y}$ are independent, then the joint distribution $P_{(\mathcal{X},\mathcal{Y})}$ will be identical to the product of the marginal distributions, implying that the mutual information is zero in given equation \eqref{infogain}.
\vspace{0.6em}
\begin{equation}\label{infogain}
I(\mathcal{X};Y) = \sum_{y \in \mathcal{Y}} \sum_{\mathrm{x} \in \mathcal{X}} p_{(\mathcal{X},\mathcal{Y})}(\mathrm{x},\mathrm{y}) 
log\Big(\frac{p_{(\mathcal{X},\mathcal{Y})}(\mathrm{x},\mathrm{y})}{p_{\mathcal{X}}(\mathrm{x}) p_{\mathcal{Y}}(\mathrm{y})}\Big)
\end{equation}

This provides a measure of how much uncertainty is reduced for one random variable by knowing information about the other. Mutual information can also be formulated as an expectation value of the KL Divergence, as shown in equation \eqref{expdkl}.
\vspace{0.6em}
\begin{equation}\label{expdkl}
I(\mathcal{X};\mathcal{Y}) = \mathbb{E}_{\mathcal{Y}} \big[ D_{KL}(p_{\mathcal{X}|\mathcal{Y}} || p_{\mathcal{X}}) \big]
\end{equation}

\vspace{0.8em}
\section{Temporal Multimodal Learning}
\subsection{Learning urban traffic} 
Daily commutes can be unexpectedly protracted by road closures, accidents, and inclement weather. The quick restoration of traffic flow through the coordinated responses of emergency vehicles may help alleviate the traffic delays impacting the road network \cite{darko2021distributed, pdroneTIM}. However, the delays, which can exceed users’ planned commuting time, can cause missed meetings, canceled appointments, and child care fees, accumulating costs. The majority of users react similarly to the unforeseen traffic delays and may unknowingly, collectively transfer congestion from one route to another \cite{BenAkiva1999}. Current navigation systems (e.g., Google Maps) are not customized to users’ tolerance for unexpected delays therefore they cannot predict optimal routes \cite{Macfarlane}. Because the network is dynamic, the route suggestion users receive at the outset of their commute may not be optimal when they are on the road \cite{Claes11}. In the literature, other traffic sensing technologies commonly fail to provide network-scale predictions under unexpected conditions. For example, current dynamic route choice models \cite{gendreau2015time} consider that the link travel time realization is only based on nearby links. The multimodal multivariate uncertainty caused by unobserved varying traffic patterns through the day has not been considered. 

To accommodate this, recent Google Deep-mind research has been using many factors and real-time updates of traffic data for more accurate prediction of travel time. Anticipatory routing guidance \cite{Hegyi19} is effective in knowledge transfer, however, ignores the potential information gain from probability density functions with \textit{more than one peak}. Consider a network with a grid laid on top in Figure \ref{fig:teaser}, where each cell represents a small geographical region. To find an optimal route from an origin cell to a destination, forecasting the condition of intermediate cells is critical. Routing literature  \cite{darko2022adaptive} did not use a location's observed data to forecast conditions at distant non-contiguous locations' unobserved data. We aggregate the data from all cells in the grid and cluster cells that have similar combinations of probability distributions. When one cell of a cluster is explored, the information gained from the explored cell can partially remove uncertainty about the conditions in distant non-contiguous unexplored cells of the same cluster. 

\begin{figure}[!htb]
	\centering
	\includegraphics[width=\linewidth]{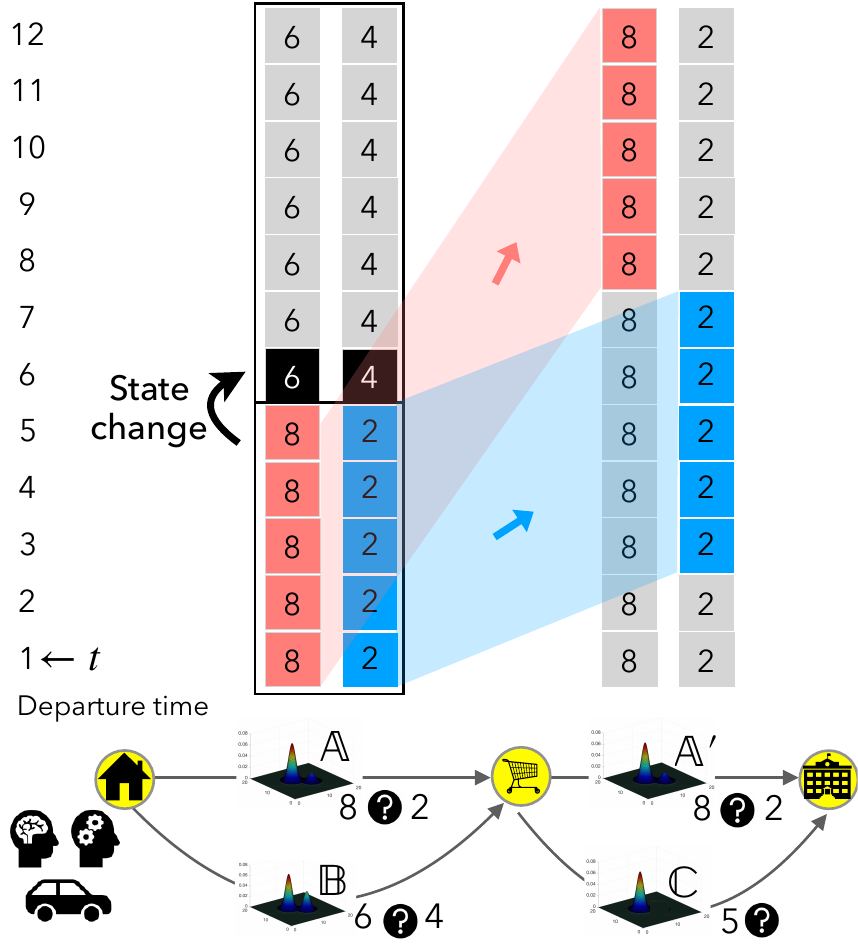}\vspace{-0.0in}
	\caption{Temporal Multimodal Learning (TML) from correlation of time-varying bimodal distributions between  links. }
	\label{uncertainty}\vspace{-1.5em}
\end{figure}

\textbf{Multimodal traffic learning.} Assume we know a freeway link $\mathbb{A}$ historically takes 2-minutes without congestion but it may take 8-minutes due to an unexpected event (e.g., incidents). We can cluster $\mathbb{A}$ and $\mathbb{A}'$ in the same correlated group assuming the bimodal travel distributions for both links are similar. Literature ignores three benefits of sending a platoon of vehicles to $\mathbb{A}$ instead of $\mathbb{B}$ shown at the bottom in Figure \ref{uncertainty}: For a scenario that turned out to be 2-minutes due to the fast clearance of the incident, 1) we can update the predicted travel time on this link $\mathbb{A}$ so other drivers can switch either their departure time or route to take this 2-minutes shortcut, 2) we can update travel time on other links (e.g., $\mathbb{A}'$) having the same type of probability distributions. By knowing that the total travel time of a route is 4-minutes, we can send more vehicles to this route and relieve other route congestion that turned out to be 8-minutes due to the long clearance of the incident, 3) we update travel time on other links (e.g., $\mathbb{A}'$) having the same type of probability distributions. By knowing that the total travel time of a route $\mathbb{A}\mathbb{A}'$ is 16-minutes, we can inform fewer vehicles to use this route, redistribute traffic to other routes (i.e., $\mathbb{B}\mathbb{C}$) having shorter travel times. While the current routing literature realize only nearby links, the realization of multimodal travel time distributions that are derived from real-world data have not been studied. 

However, recent studies \cite{Guo10, Zheng10} have shown that travel time distributions on freeways have two or more modes as distinct peaks in the probability density function due to the mixes of driving patterns and vehicle types. This multimodal (or bimodal) distribution exists on arterial roads, where a vehicle passing a signal at the end of the green would experience quite different travel time than the vehicle following behind it that must make a stop at the red, although they traveled next to each other. Without knowing the future traffic with confidence, the traditional choice theory considers the bounded rationality \cite{Han06, HAN201516, GUO2013203, DI201674} of the majority of agents taking a detour to link $\mathbb{B}$, which causes congestion on $\mathbb{B}$ and nearby roads. 

\textbf{Temporal multimodal traffic learning.} In Figure \ref{uncertainty}, $\mathbb{A}$ has a bimodal distribution with a mode at 8 and 2 between time stages 1 to 5, \textit{switching} to a bimodal distribution with a mode at 6 and 4 at time stage 6. For departing at time stage 1, the time-invariant method adds travel time together either the high or low modes of link and route $\mathbb{AA'}$ travel time to be either 8+8 or 2+2. The time-variant method accounts for the time needed to traverse $\mathbb{A}$ either \textcolor{red}{8} or \textcolor{blue}{2} minutes and re-evaluates the travel time at $\mathbb{A'}$ based on the time of entering $\mathbb{A'}$, may encourage a detour to Link $\mathbb{C}$ in case of 8 minutes of realization. We assume that the state change is given based on event models \cite{park16c}.

The previous examples assume that the primary factor contributing to travel time variation on a given link is the time of day. The focus of this study is to use travel time correlation information to remove uncertainty in within-day travel. Travel time may depend on other factors such as day-of-week, weather patterns during the day, and special events in the region like post-game-day traffic near a sports stadium. If data on these other variables are made available, the same temporal learning process can be extended (see Section 6 for temporal learning under the presence of multiple variables of interest). 


\vspace{-0.5em}
\subsection{Improving KF prediction}
Prediction uncertainty in travel time is improved by considering TML on real-world traffic data. Let $C$ be the set of all links, traffic message channels (TMCs), across the network and $T$ be a finite set of discrete-time intervals over the morning peak period \cite{park15c, park16c}. We consider 39 TMCs ($|C|=39$) on Interstate 540 in Raleigh, NC during 24 ten-minute time intervals from 8:00 am to 12 noon ($|T|=24$). Probe-vehicle-based speed for each TMC was obtained from the National Performance Management Research Data Set (NPRMDS).

NPRMDS contains the travel time and speed information for each TMC for each time interval across different days over the course of eight months. Due to the day-to-day traffic randomness, the traffic speed on TMC $c\in C$ for time interval $t\in T$, denoted by $v_{c}^{t}$, is a random variable. As argued in the literature \cite{PARK2010662, Guo10, Zheng10}, $v_c^t$ is likely to have a probability distribution with multiple modes (multimodal distribution). We learn and predict $v_c^t$ within day by analyzing the spatiotemporal correlations between random variables $v_c^t$ for all $(c,t)\in C\times T$. By clustering all $v_c^t$ variables, we identify spatiotemporal patterns and different combinations of traffic speed distributions with following steps: 
\begin{itemize}[noitemsep, topsep=0pt]
	\item Analyze the spatiotemporal probability distribution of variables $v_c^t$ by aggregating variation of traffic speed for a specific time interval across eight months. For this case study, we assume that the only factors influencing travel time are the location of TMC ($c$) and time-of-day interval ($t$). 
	\item Clustering is performed across all $C$ and $T$ using minimum message length criteria to identify TMC's with similar probability distributions \cite{SNOB}. Clustering algorithm will automatically discover the optimal number of clusters.  
\end{itemize}

\begin{figure}
    \centering
    \includegraphics[width=\linewidth]{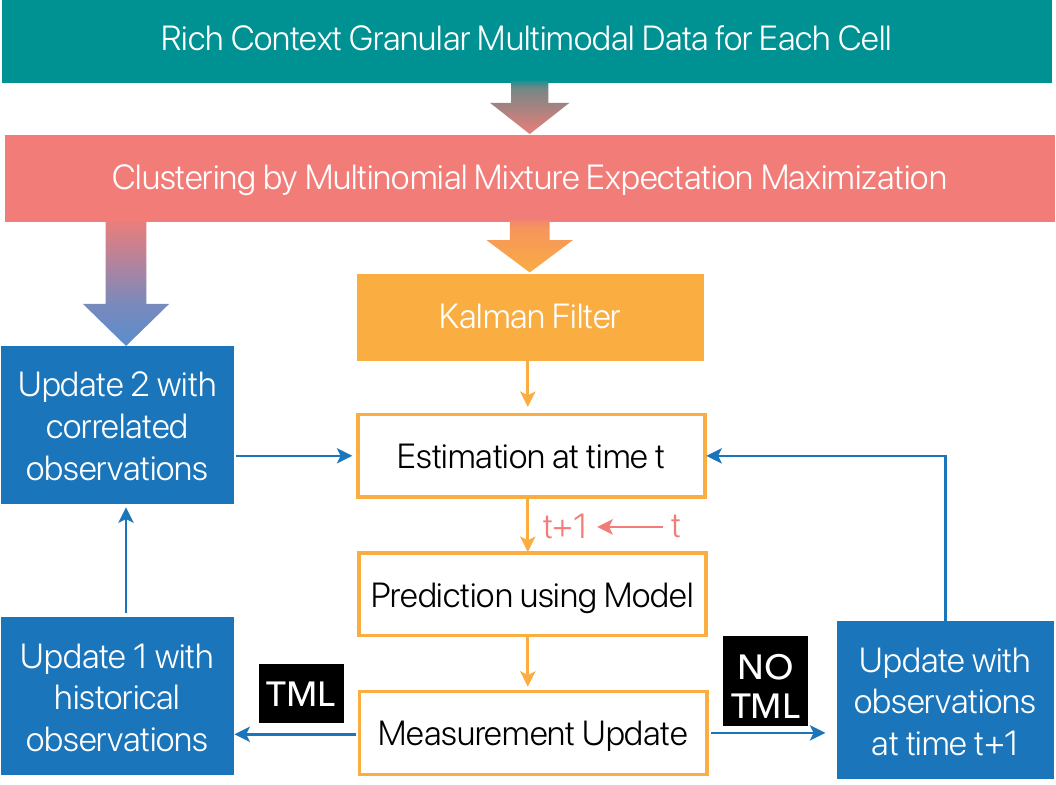}
    \caption{Two steps in KF-TML: In the predict step, a model is employed to predict the chosen state variable at next time interval $t+1$ using measurement from previous time interval ($k$). In the update step, the predicted state is corrected using the noisy measurements at $t+1$.}
    \label{KF}\vspace{-1.5em}
\end{figure}

\textbf{Kalman Filtering (KF) Prediction $v_c^t$.} We model the evolution of random variable $v_c^t$ from one 10 minute interval to the next interval within-day with and without information gain using KF. 
The data of $v_c^t$ acquired from TMCs have inherent noise due to sensor errors. Employing KF can produce an accurate estimate of $v_c^t$ using noisy measurements over the period (24 intervals of time). In this paper, the traditional KF is expanded to consider the information gain from the clustering step.
We model evolution of variable $v_c^t$ from the first time interval $t=8-8:10am$ to the last interval $t=11:50-12pm$ within a day. Figure \ref{KF} shows the KF process which is formulated in the following equations.

\textbf{Prediction step.} Projection of the state at time $t$ using the prediction at previous time $t-1$ is given by:
\begin{equation}
\vspace{-1.0em}
    \hat{x}^{-}_{t}=A\hat{x}_{t-1}^{+}+B\mu_{t} 
\end{equation}
where,
\begin{itemize}[noitemsep, topsep=0pt]
    \item $\hat{x}_{t-1}^{+}$ is the state vector of the process at time $t-1$. 
       In this case, state vector considered is
$\begin{bmatrix}
    \text{speed}\\
    \text{acceleration}
\end{bmatrix}$, 
     where, acceleration is defined as the rate of change of speed of TMC with respect to previous time period.
     \item Matrix $A$ is the state transition matrix of the process from the state at $t-1$ to state at $t$ and is assumed stationary over time. That is, 
     $A=\begin{bmatrix}
         1 & dt \\
         0 & 1
     \end{bmatrix}$
     \item $dt=1$ according to definition of acceleration defined above.
    \item Matrix $B$ is a matrix of all zeros as there is no known external control input factor that affects speed measurement.
    \item $P$ is the error covariance matrix. It is interpreted as the error in estimation according to filter.
    \item $Q$ is the process noise defined as
        $Q=\begin{bmatrix}
         0.04 & 0 \\
         0 & 1
     \end{bmatrix}$
    \item We assumed the speed with a variance of 0.04 in prediction step. 
\end{itemize}   
 
Projection of error covariance of state
\begin{equation}
    P^{-}_{t}=P_{t-1}^{+}A^{T}+Q
\end{equation}
\textbf{Correction step.} In this step, we determine the Kalman Gain at time $t$ (denoted by $K_t$) which can be interpreted as,

\begin{equation}
    \text{Kalman gain}=\frac{\text{Uncertainty in prediction}}{\text{Uncertainty in prediction + measurements}}.
\end{equation}

We can write,
\begin{equation}
    K_{t}=P^{-}_{t}H^{T}(HP_{t}^{-}H^{T}+R)^{-1}
\end{equation}

where, $H$ is the connection matrix between the state vector and the measurement vector and $R$ is the data precision matrix. In our case, 
    $H=\begin{bmatrix}
         1 & 0 
    \end{bmatrix}.$ 

In the next step of KF, speed prediction is updated using observations $Z_{t}$. In case of KF-no TML, $Z_{t}$ are the speed observations on a given day while in case of KF-TML, $Z_{t}$ are mean and variance of historical speed data.   
\begin{equation}
    \hat{x}_{t}^{+}=\hat{x}_{t}^{-}+K_{t}(Z_{t}-H\hat{x}_{t}^{-})
\end{equation}
Error covariance matrix is also updated in this step using the Kalman gain.
\begin{equation}
    P_{t}^{+}=(I-K_{t}H)P_{t}^{-}
\end{equation}

KF-TML has an additional step as the speed prediction update with data $Z_{t}^{+}$ obtained from information gain of correlated links. 
\begin{equation}
    \hat{x}_{t}^{++}=\hat{x}_{t}^{+}+K_{t}^{+}(Z_{t}^{+}-H\hat{x}_{t}^{+})
\end{equation}
This step also updates the error covariance matrix.
\begin{equation}
    P_{t}^{++}=(I-K_{t}^{+}H)P_{t}^{+}
\end{equation}

The hat operator indicates an estimate of a variable. The superscripts -, + and ++ denote predicted (prior), updated 1 (posterior 1) and updated 2 (posterior 2) estimates, respectively.  The posterior 1 will be the final prediction in KF-no TML while posterior 2 will be final outcome in KF-TML. 

During the update step, observations available from the correlated links from previous time intervals are considered. The mean and variance of speeds of all correlated links are used as the new observation in the update step. Therefore, traditional KF has only one update step but in this paper, the algorithm is modified to have two updates, one with mean and variance of historical data of 8 months and the other with mean and variance of correlated speed data obtained from the clustering step.  

The $(1,1)$ element in matrix P denotes the variance in estimation of speed. Percentage change in $P(1,1)$ with information gain with respect to $P(1,1)$ without information gain is calculated.

\begin{equation}
    \Delta P =  \frac{P(1,1)_{\text{without info gain}}-P(1,1)_{\text{with info gain}}}{P(1,1)_{\text{without info gain}}}*100
\end{equation}

\begin{figure}[!htb]
	\centering
	\includegraphics[width=\linewidth]{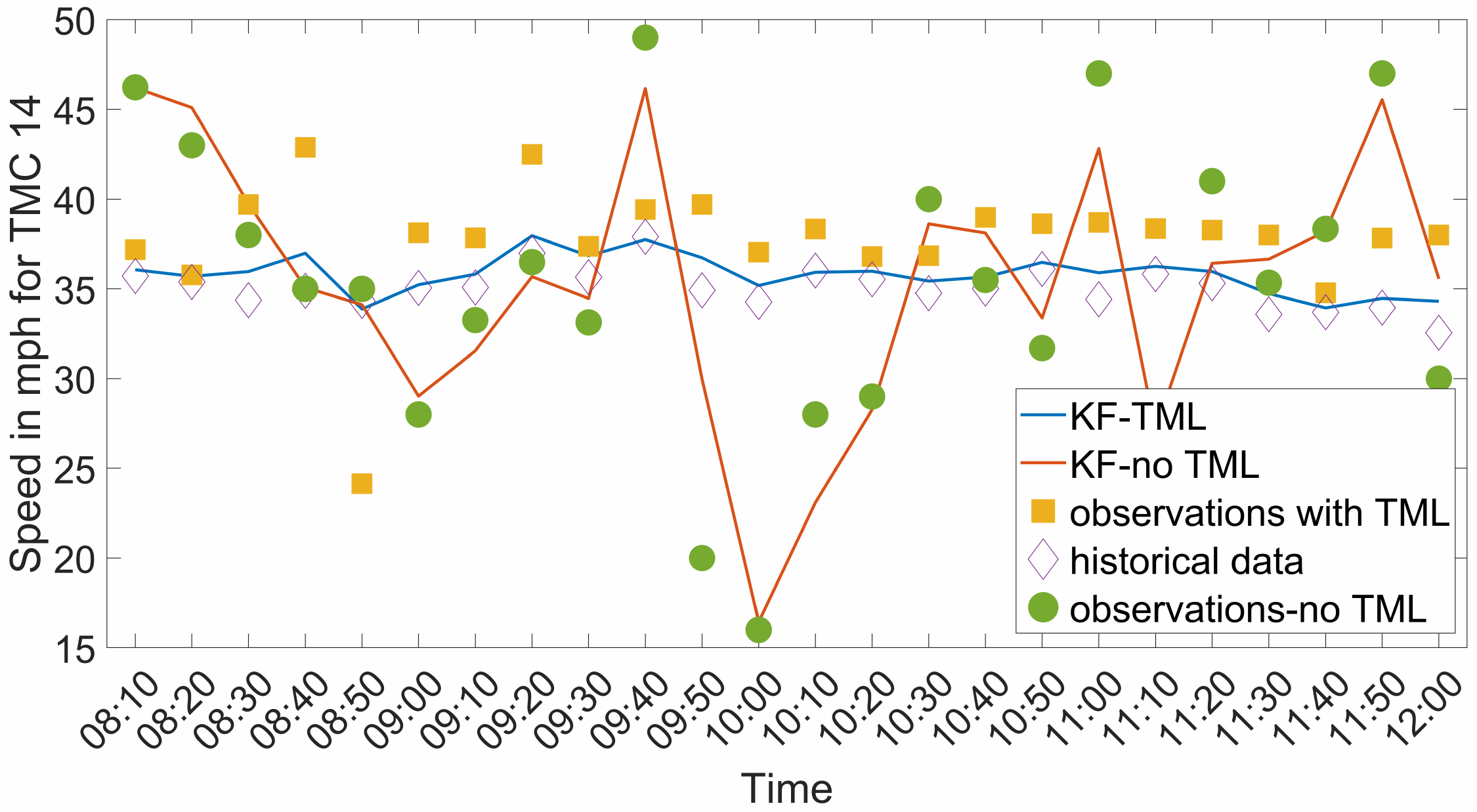}\vspace{-0.0in}\vspace{-1.0em}
	\caption{Speed predictions with and without TML and corresponding observations}
	\label{TML}\vspace{-0.2in}
\end{figure}

\textbf{Results.} The performance of KF with TML is compared against the benchmark. Traditional KF without TML ignores the correlation information where the observation is simply the observed speed from the sensor on a given day, and KF with TML is modified to include the mean observation of speed from other TMCs and previous time-periods that are within the same cluster as the given TMC and time-period. Figure \ref{TML} shows that the KF prediction with TML has fewer errors compared to the KF prediction without TML. Figure \ref{TML1} in Appendix \ref{appendixc} shows the percentage change in uncertainty of predictions when TML is considered. A significant reduction in uncertainty indicates more confidence in the predictions with TML.  In KF without TML, the update step uses measurements with noise at time $t$ to get accurate predictions at time step $t$. In KF with TML, we improve the prediction performance of traditional KF by using the correlated observations from previous periods and it helps to achieve the estimation of speed at $t$ on the previous time step, $t-1$. This improved method is useful in getting more accurate predictions ahead of time.     



\section{Temporal Multivariate Multimodal Learning }
\subsection{Learning storm atmosphere} 
With observations from sUAS, NOAA's National Hurricane Center can better measure critical variables and parameters in the boundary layers of hurricanes \cite{cione2016coyote}. The accuracy of the data collected by the sUAS agreed well with that of the manned measurement, with the sUAS sometimes capturing more variability than the manned measurement \cite{cione2019eye}. However, the predefined navigation procedures do not necessarily consider how data gathered from a flight path improves the hurricane forecasting. The criteria for location selection was ``\textit{difficult to observe in sufficient detail by remote sensing}''. We find the ``\textit{optimal routes considering importance of observations}'' gained from the data in a precise target location among high-dimensional spaces. We analyze how  online updating from sUAS collected meteorological data would benefit hurricane intensity forecasting considering the temporal variation in the uncertainty of hurricane prediction. The temporal multivariate learning and in-situ data collections can significantly improve understanding of hurricane movement, relevant dynamics, and track prediction with the same effort and less risk. 

\begin{figure}[!htb]
    \centering
    \includegraphics[width=\linewidth]{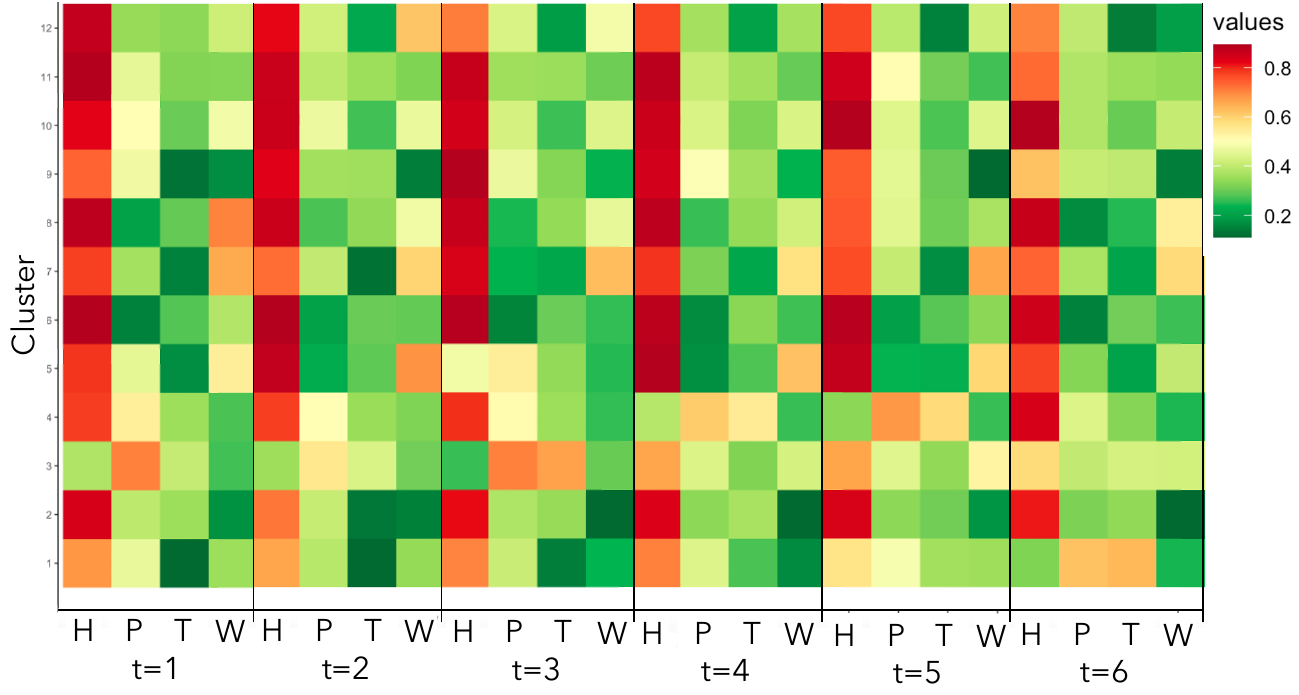}\vspace{-0.1in}
    \caption{The clustered temporal multivariate distribution of four variables across all time stages (4 $\times$ 12 dimensional). For example, one observation of temperature within cluster 1 at time stage 1 are used in removing prediction uncertainties of humidity in cluster 2 at time stage 3.}
    \label{fig:tmmlcluster}
\end{figure}

\textbf{TMML in hurricane forecasting.} To generate the uncertainty distribution for Hurricane Harvey, conventional in situ observations (e.g., Dropsondes) and all-sky satellite radiance from GOES-16 were assimilated in a state-of-the-art data assimilation system (ensemble KF) and built around the Advanced Weather Research and Forecasting Model (WRF-ARW) and the Community Radiative Transfer Model (CRTM) to provide hourly temporal resolution forecast of Hurricanes \cite{Masashi2020}. Indirect learning will overcome possible limitations in observing only one reliable sample variable in a cell out of many sensor payloads \cite{poterjoy2011dynamics}. The temporal multivariate cluster groups cells with a similar combination of distribution of multiple variables in each cell: e.g., temperature (T), pressure (P), wind speed(W), relative humidity (RH)  (Figure \ref{fig:tmmlcluster}). Once we have an observation of one variable, we update posterior of other variables at the same location/other locations at the same time/other time stages with the same cluster as the observed location. 

\subsection{Improving sUAS routing}
\begin{figure}[!htb]
	\centering
	\includegraphics[width=\linewidth]{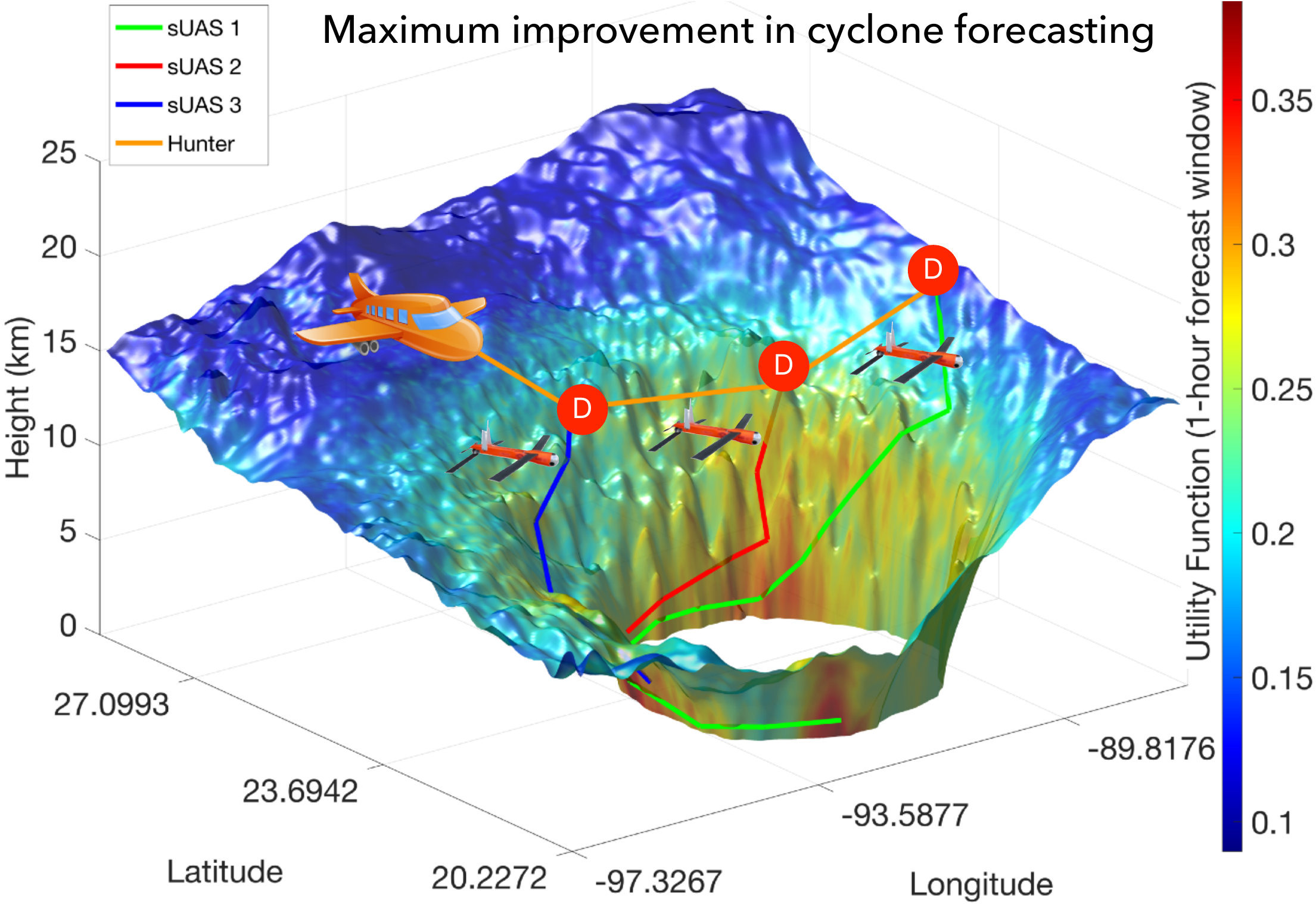}\vspace{-0.0in}
	\caption{Active sUAS In-situ Sensing}
	\label{4d_rep}
\end{figure}

Once the first sUAS is launched from a tube attached to a P-3 hurricane hunter aircraft and controlled remotely from the airplane to be deployed to the lower layer, the benefit of the online update of the information presents an increase in accuracy relative to the initial hurricane prediction (Figure \ref{4d_rep}). The overall improvement in prediction uncertainty after observations are made along a path is computed as the difference between the sum of measurement variances before and after observations.  

\begin{figure*}[!htbp]
	\centering
	\vspace{-0.00in}{
		\includegraphics[width=\linewidth]{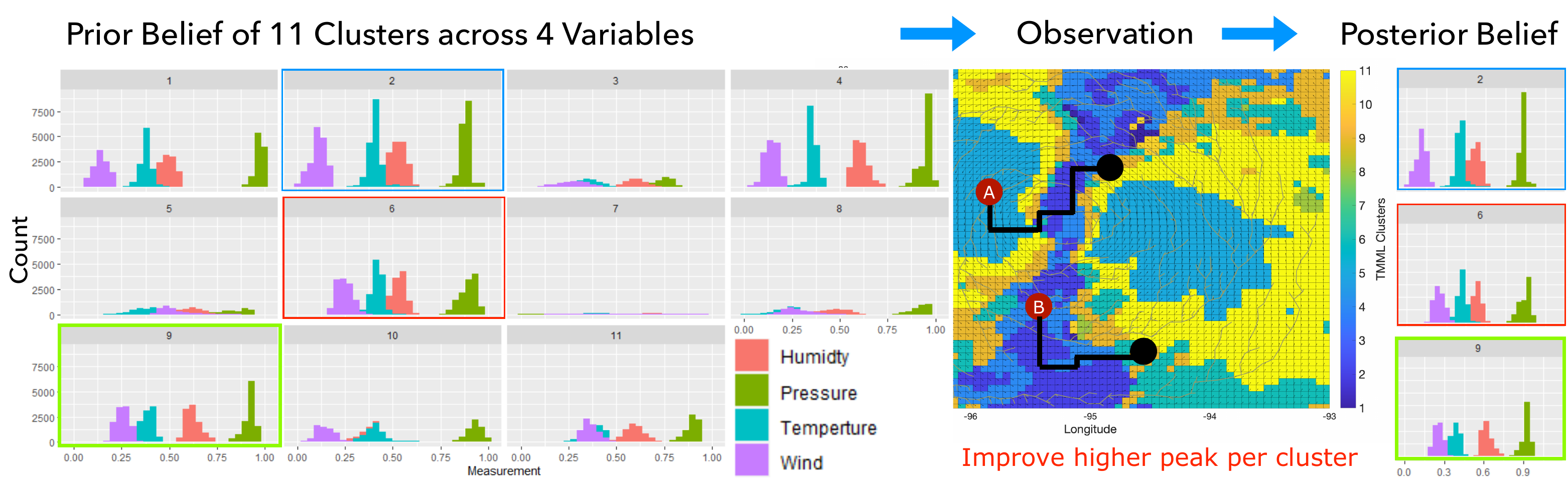}\vspace{-0.1in}}
	\vspace{-0.1in}\caption{Posterior update after agents A \& B depart from red and observe 64 cells out of 2500. As the probability distribution for each variable for each cluster in the plot is narrower, it represents less uncertainty. Variational Bayesian inference generates a posterior belief given the prior belief of cell-type distributions. As observations made along the route, particularly the cluster 2, 6, and 9 observations are made directly and indirectly would be further reduced uncertainty and present narrower probability distribution.}
	\label{hinto}\vspace{-0.5em}
\end{figure*}

\begin{figure}[!htbp]
	\centering
{\includegraphics[width=\linewidth]{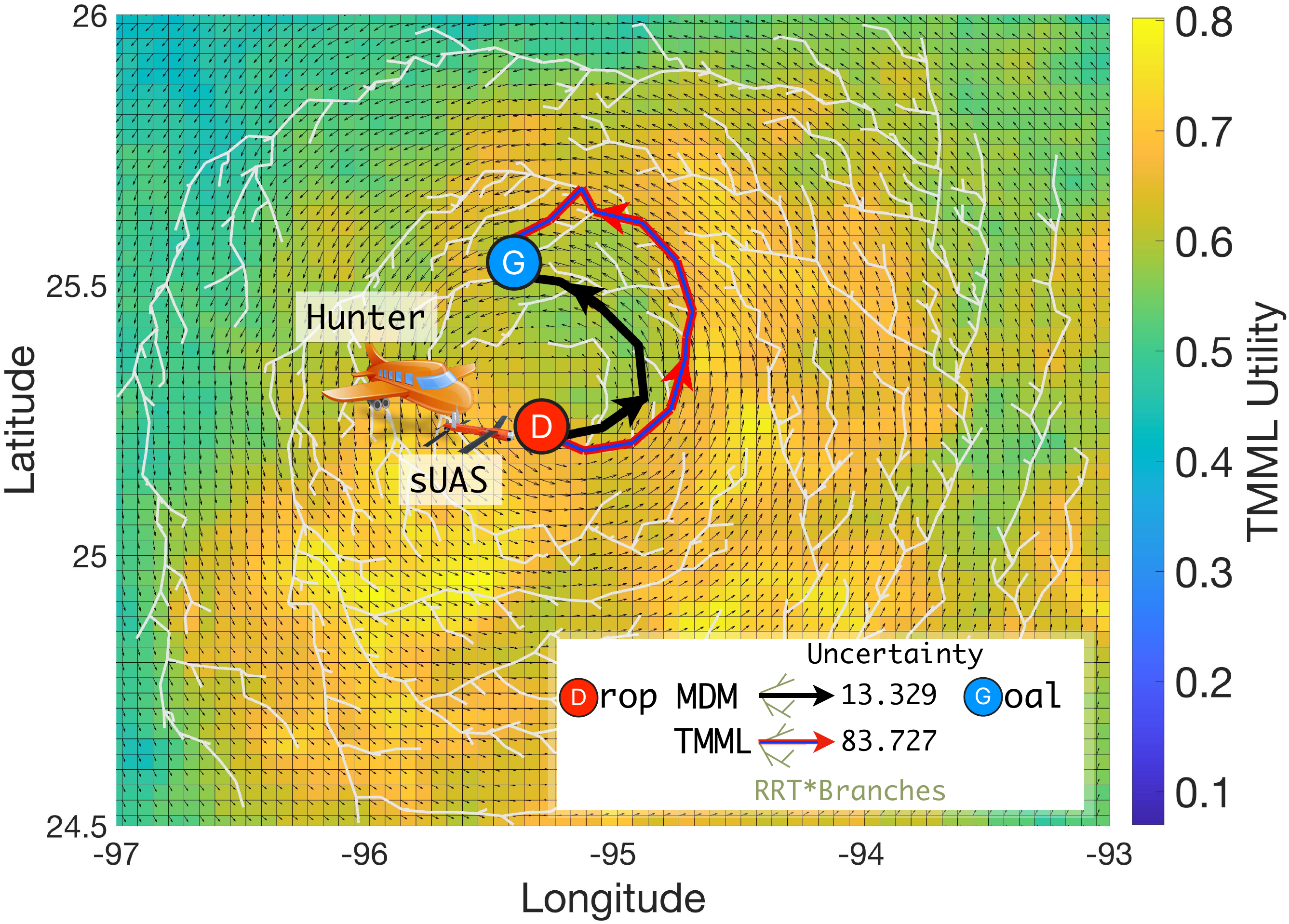}\vspace{-0.2in}	}
	\caption{Safe and efficient sUAS path solution by TMML-RRT* and MDM on centered horizontal cross-sections for Hurricane Harvey at a 1.1-km level.}
	\label{prehurr}\vspace{-0.2in}
\end{figure}
The utility map for the agent-guided observations of the hurricane environment is constituted from the standard deviation and multivariate cluster entropy between the four state variables: pressure, temperature, wind speed, and relative humidity. These components are normalized to ensure a consistent scale. First, the normalized standard deviation $\hat{\sigma}_{(x,j)}$ of the measurement for these state variables $x \in X$ are linearly combined as \begin{math}\sum_{x \in X} \hat{\sigma}_{(x,j)} \end{math} to quantify the overall standard deviation $\sigma_{(j)}$ in each cell of the hurricane space. Second, the entropy $H(x, n_{type})$ for each multivariate cluster distribution
is computed as
\begin{equation}
H(\textbf{x}, n_{type}) = -\mathop{\sum_{t=1}^{T}\sum_{k=1}^{K}} P\left(\textbf{x},n_{type}(k),t\right) \log _{2} P\left(\textbf{x}, n_{type}(k),t\right)
\end{equation}
where $ P\left(\textbf{x},n_{type}(k),t\right) $ is probability of category $k\in K$ in cell $j$ of cluster $ n_{type}$ at time $t$ and $\textbf{x}$ is an $N$ dimensional vector of the state variables.
For illustration, in a 12-hour forecast window for Hurricane Harvey, cells with similar characteristics were grouped into 11 clusters as different combinations with multiple variables, where the optimal number of clusters and distribution parameters in each cluster were estimated to maximize posterior probability (Figure \ref{hinto}). Variational Bayesian inference approximates the true posterior based on the $\mathbb{K}\mathbb{L}$ divergence minimizing the information lost with optimal variational parameters. In Appendix \ref{appendixa}, the estimation of expected observation and posterior estimate of the variances after observations by the sUAS is presented.    

\textbf{Results.} Although unobserved non-contiguous cells may not share any inherent correlation with locally observed cells, classification errors are found be correlated with certain features found in different locations. By clustering cells with similar distributions (e.g., predicted wind speed) and correcting those errors as more evidence becomes available from sUAS observations, the reliability of the prediction was greatly improved (Figure \ref{prehurr}). Anticipatory sUAS routing lowered the overall energy usage, and maximized the reduction of forecasting error by exploring and sampling unobserved cells along the path to a target location. This new method significantly improved the combined quality of observations (pressure, temperature, and wind speed) against the Minimum Distance Method (MDM) in Hurricane Harvey  \cite{parkESS2020}. In this paper, the multistage online decisions are made by combining the benefits of direct sensing through the sensitivity map and the additional benefits from indirect learning through the correlation structure considering three components: 1) temporal learning, 2) multimodal learning from one observed geographical location to other similar locations, 3) deep multivariate learning by grouping similar covariance of all four variables rather than estimating correlation of each pair. Compared against unimodal univariate learning, the combination of all three components presents a larger reduction in prediction uncertainty. 
\vspace{-1.0em}
\begin{table}[!htbp]
	\caption{Improvement in Prediction Uncertainty}\vspace{-0.5em}
	\label{tab:summ}
\begin{tabular}{llcccc}
\toprule
\multicolumn{2}{c}{\multirow{2}{*}{MDM = 2.36\%}} & \multicolumn{2}{c}{Non temporal} & \multicolumn{2}{c}{\multirow{2}{*}{Temp.}} \\\cmidrule(lr){3-4}

\multicolumn{2}{c}{}                  & Unimodal          &   Multimodal       & \multicolumn{2}{c}{}                  \\\midrule
\multirow{2}{*}{Correlation}          &  Univariate        &    2.66\%       &   4.34\%       & \multicolumn{2}{c}{-}                  \\
                           &    Multivariate      &   3.28\%        &  6.25\%        &  \multicolumn{2}{c}{11.26\%}                  \\
\multicolumn{2}{c}{Deep multivariate}           &- & 7.37\%& \multicolumn{2}{c}{13.45\%} \\              \bottomrule  
\end{tabular}\vspace{-0.5em}
\end{table}

The average improvement in predicted multimodal measurement was significantly higher when TMML were considered. MDM \cite{parkESS2020} ignores those two properties by averaging to reduce the dimension of the data, but these results show that while the dimension may have increased, uncertainty reduction in temporal correlation may have reduced the size of the deterministic problem.

\section{Conclusion}
With the Temporal Multimodal Multivariate Learning (TMML), we have introduced a new family of RL models that can indirectly learn and transfer information from multiple modes of probability distributions of multiple data variables in different time stages. These models can solve challenging tasks where the uncertainty is revealed in a sequence by grouping samples within similar distribution types and inferring the posterior based on expected observations. The effectiveness of TMML has been demonstrated on real-world autonomous navigation in urban transportation and Hurricane. TMML opens appealing research opportunities in the study of information-theoretic decision making that exhibit nontrivial indirect learning from spatiotemporal correlation.

\section{Acknowledgments}
Part of the research work was carried out at the NASA Jet Propulsion Laboratory (JPL), California Institute of Technology, under a contract with the National Aeronautics and Space Administration (NASA) [80NM0018D0004]. Funding for this research was provided by NSF [1910397, 2106989], NASA JPL [RSA 1625294, 1646362, 1659540], and NCDOT [TCE2020-01]. The authors thank JPL Education Office for providing internships and faculty visiting opportunities to conduct this research and Dr. Joe Cione for giving us valuable insight into past, current, and future sUAS missions and key ﬁndings. The data used for these analyses are available at: \url{https://figshare.com/s/90f31f60e5821dae90bd}

\bibliographystyle{ACM-Reference-Format}
\bibliography{kdd2022}

\appendix

\section{Posterior approximation}\label{appendixa}
We combine observations and prior data with the importance of information. The sequence of observations are assimilated with the prior predicted measurements to provide the best estimate (posterior) of the measurements. The multivariate measurements are represented as grid-point values with background prior information at location $i$: 

\begin{equation}
M_{\textnormal{Pred}(\textbf{x}, i)} + \sigma_{\textnormal{Pred}(\textbf{x}, i)}^2,
\end{equation}

and after observation at location $i$:

\begin{equation}
M_{\textnormal{Obs}(\textbf{x}, i)} + \sigma_{\textnormal{Obs}(\textbf{x},i) }^2,
\end{equation}

\noindent where $M_{\textnormal{Pred}(\textbf{x}, i)}$ is the mean predicted measurement at location $i$, $\sigma_{\textnormal{Pred}(\textbf{x}, i)}^2$ is the variance of predicted measurement at location $i$, $M_{\textnormal{Obs}(\textbf{x}, i)}$ is the sUAS observation at location $i$, $\sigma_{\textnormal{Obs}(\textbf{x},i)}^2$ variance of sUAS observation at location $i$.
The variance $\sigma_{\textnormal{Obs}(\textbf{x},i) }^2$ represents the imperfections of observations made by sUAS sensors. Our model assumes that this variance is known apriori based on the type of sensors used. The best estimate of the measurement of variable $x$ at location $i$ is written as:
\begin{equation}
M_{\textbf{best estimate}(\textbf{x}, i)} = (1-\beta) M_{\textnormal{Pred} (x, i)} + \beta M_{\textnormal{Obs} (\textbf{x}, i)}.
\end{equation}

\noindent $\beta$ is the weight between the predicted measurements and observation. The best estimate of weight considers the variance of the predicted measurement and observation, written as:

\begin{equation}
\beta = \frac{\sigma_{\textnormal{Pred} (\textbf{x},i)}^2}{\sigma_{\textnormal{Pred} (\textbf{x},i)}^2 + \sigma_{\textnormal{Obs} (\textbf{x},i)}^2}.
	\label{eq:binst}
\end{equation}

\noindent The variance of the best estimate of measurement for variable $\textbf{x}$ at location $i$ is less than that of either the prediction or the observation:

\begin{equation}
\sigma_{\textbf{best estimate} (\textbf{x}, i)}^2 = (1-\beta) \sigma_{\textnormal{Pred} (\textbf{x}, i)}^2.
\end{equation}

\noindent To account for the  effect of an influence region around each observation point, we introduce a weighting function $\boldsymbol{\omega (i,j)}$, to update the best estimates of the variance at each grid locations $j$ in the vicinity of observation point $i$ written as: 

	\begin{equation}
	\boldsymbol{\omega (i,j)}=\max \left(0, \frac{R^{2}-d_{i, j}^{2}}{R^{2}+d_{i, j}^{2}}\right)
	\end{equation}

\noindent where $d_{i,j}$ is a measure of the distance between points $i$ and $j$. The weighting function $\boldsymbol{\omega (i,j)}$ equals to one if the grid point $j$ is collocated with observation $i$. It is a deceasing function of distance which is zero if $d_{i,j} \geq R$. $R$ (``\text{the influence region or radius}'') is a user defined constant beyond which the observations have no weight. The modified best estimate of the variance at each grid point location $j$ can now be written as:
\begin{equation}
	\sigma_{\textbf{best estimate} (\textbf{x},j)}^2 = (1-\beta* \boldsymbol{\omega (i,j)}) \sigma_{\textnormal{Pred} (\textbf{x}, j)}^2.
\end{equation}
	

Sequential learning updates a cells entropy and in extension the utilities each time an observation is made in other cells belonging to the same cluster. We introduce a weight $\omega(x,o_{n_{type}})$ (decreasing function of sample size $o_{n_{type}}$) that updates the entropy $H(x, n_{type})$ as observation of cells belonging to the same cluster type are made. Since there is high confidence in the measurement in cells belonging to clusters with low entropy, we exploit those low entropy cluster types through a few sampling of their member cells. A posterior update will be applied to the variance of all similar type cells in such scenarios.

Conversely, there is low confidence in the measurement in cells belonging to clusters with high entropy. Therefore, we explore the high entropy cluster types through a large sampling of their member cells. With enough sampling of member cells, we can reduce the entropy to a set threshold and apply a posterior update to measurements in its member cells. The sequential learning in environments learn the optimal sample size and update as more observations are obtained. Online recourse in Algorithm \ref{recourse1} shows the sequential update of the TMML-RRT* (Algorithm \ref{rrtstarett}).  

\begin{algorithm}[!htb]
	\caption{\textsf{:OnlineRecourse}}\label{recourse1}
	\begin{algorithmic}
		\STATE X $\gets$ $n_{s}$\\
		\FOR{t=1 to i=$\tau$}
		\IF{X = $n_{g}$}
		\STATE break;
		\ELSE
		\STATE $(X_{new},U_{new}) \gets$ max(X, $R_{X, X_{successor}}$)\\
		\IF{$U_{new}$ > $U_{X_{successor}}$}
		\STATE $X_{successor}$ $\gets$ $X_{new}$
		\STATE Rewire(X, $X_{successor}$, $X_{successor2}$)
		\ENDIF
		\STATE X $\gets$ $X_{successor}$ \\
		\ENDIF
		\ENDFOR
	\end{algorithmic}
\end{algorithm}

\section{KF-TML Prediction}\label{appendixc}
\begin{figure}[!htb]
	\centering
	\includegraphics[width=\linewidth]{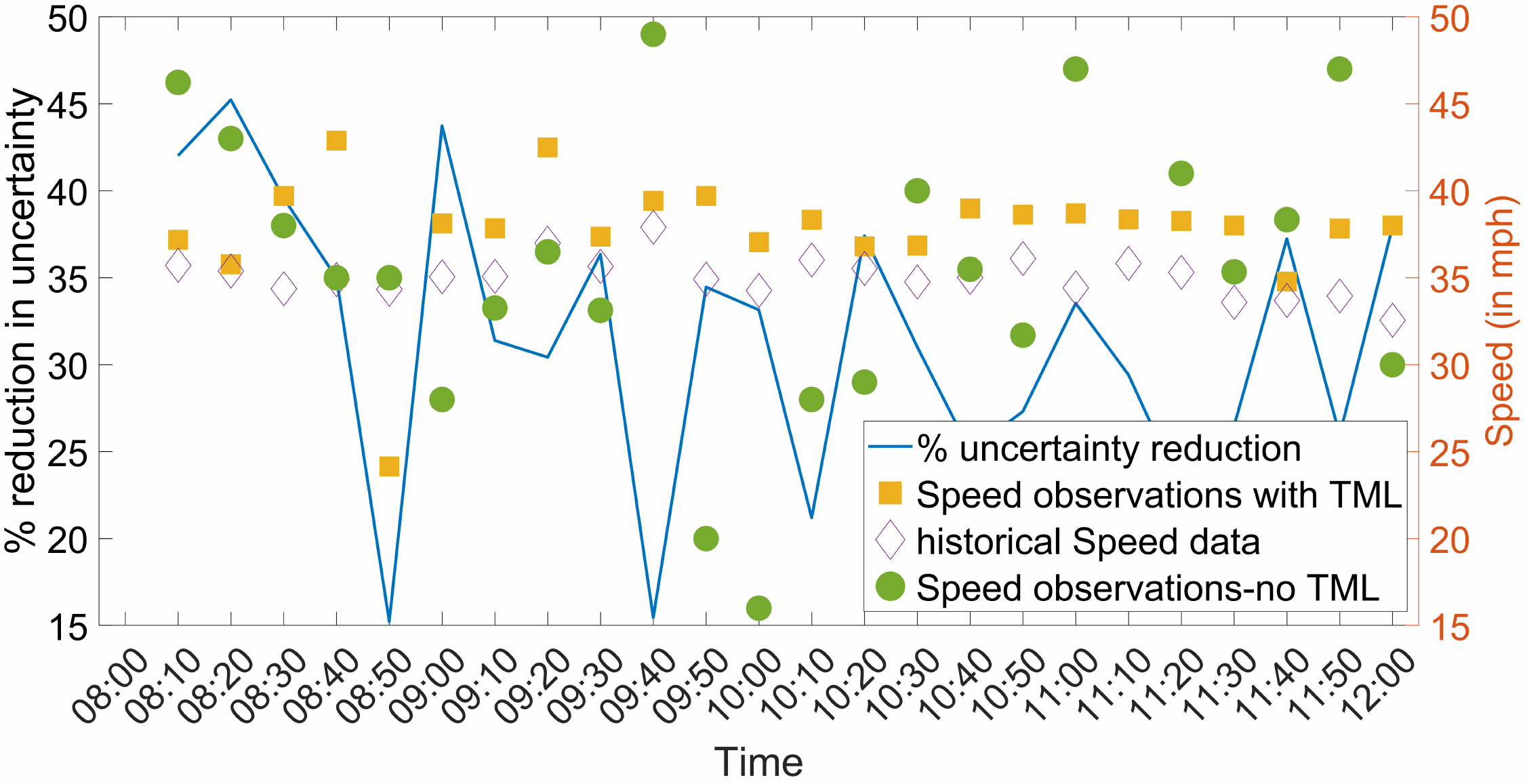}\vspace{-0.0in}
	\caption{Percent change in uncertainty of KF prediction when TML is considered}
	\label{TML1}
\end{figure}

The main goal of the KF-TML is to use the information gain from spatiotemporal correlation between TMCs to reduce uncertainty in KF speed prediction. Figure \ref{TML1} shows the significant percent reduction in uncertainty of predictions when the KF-TML is employed. When speed observations with TML are close to historic observations, the reduction in uncertainty is higher.  

\section{TMML in sUAS Routing}\label{appendixb}
\begin{figure}[H]
	\centering
	\includegraphics[width=\linewidth]{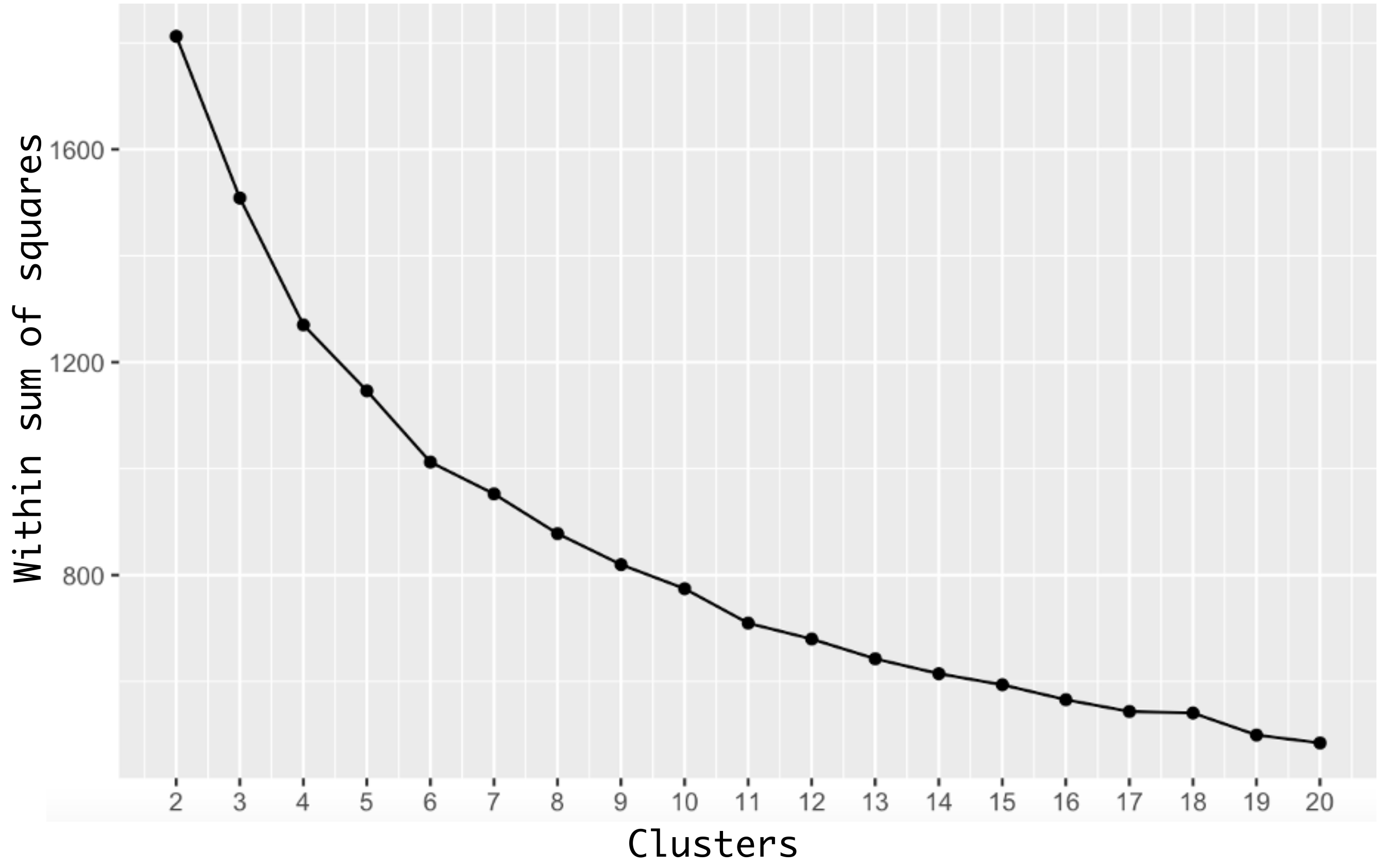}\vspace{-0.0in}
	\caption{Based on sum of squares of hurricane prediction output uncertainty, optimal cluster of 12 is used. We cluster the four factors across 12 time stages using 12 clusters to create a heap map to highlight the different clusters.}
	\label{clusters}
\end{figure}

Using k-means clustering, the map is divided into regions of similar cell types. Clusters are formed based on the entropy and the expected value of each cell. The optimal number of clusters for the dataset was calculated using a Gap function, demonstrated in Figure \ref{clusters}. Because of differences in scale, k-means clustering performs better when entropy is expressed as a percentage rather than a decimal value. 

\begin{figure}[H]
	\centering
	\includegraphics[width=\linewidth]{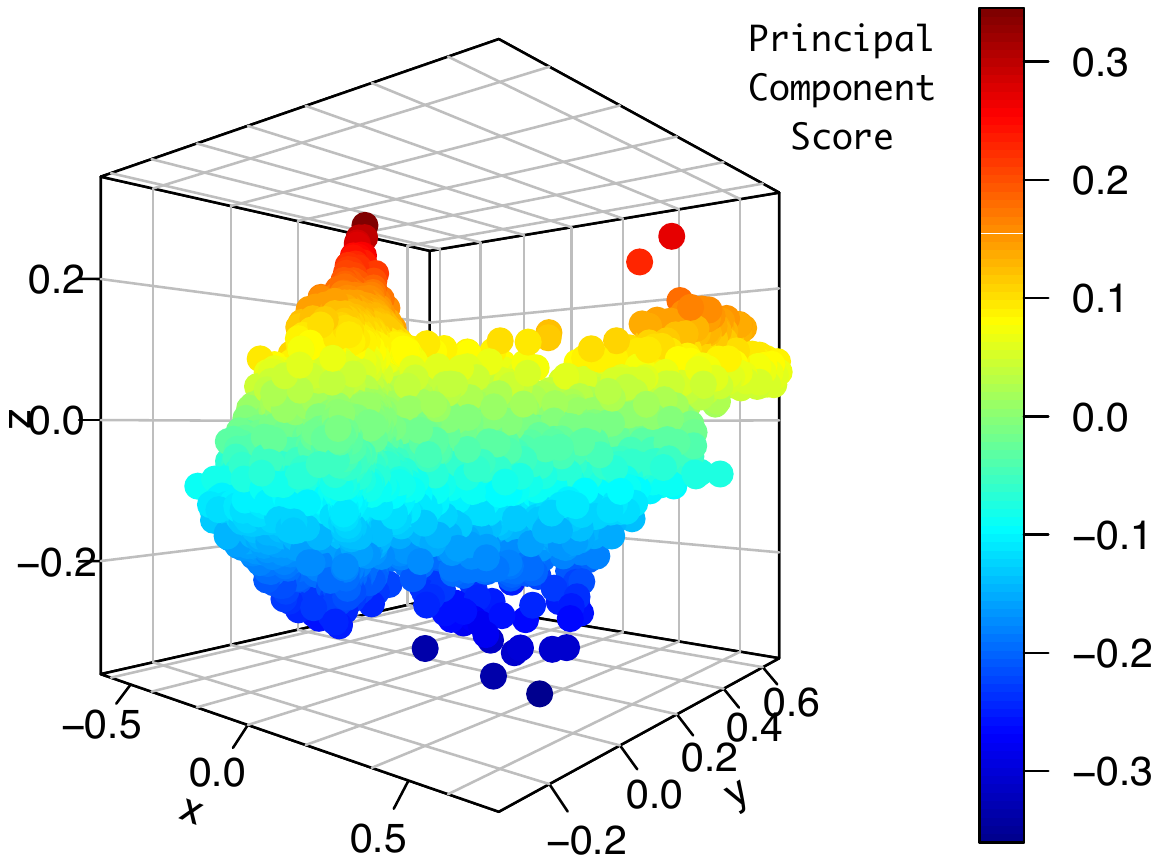}\vspace{-0.0in}
	\caption{The result of multimodal multivariate clustering and temporal multimodal multivariate is described by transforming four variables and time dimensions to three principle components in a three-dimensional graph with cumulative variation proportion of 90$\%$.}
	\label{4d_rep2}
\end{figure}

Figure \ref{4d_rep2} shows principle components of those 12 clusters for each variables across 12 hours in hurricane case study. As long as any observation is within the same cluster, based on the correlation measure, prediction uncertainty of multiple variables are updated. 

\end{document}